\documentclass[10pt,twocolumn,letterpaper]{article}

\usepackage{cvpr}              %

\usepackage[dvipsnames]{xcolor}
\makeatletter
\@namedef{ver@everyshi.sty}{}
\makeatother
\usepackage{tikz}

\usepackage{times}
\usepackage{epsfig}
\usepackage{graphicx}
\usepackage{amsmath}
\usepackage{amssymb}
\usepackage{rotating}

\usepackage{multirow}

\usepackage{upgreek}
\usepackage{booktabs}
\usepackage{mathtools}
\usepackage{physics}

\usepackage[pagebackref,breaklinks,colorlinks]{hyperref}

\usepackage{subcaption}
\usepackage[normalem]{ulem}
\usepackage[title]{appendix}

\usepackage{adjustbox}
\usepackage{array}
\newcolumntype{R}[2]{%
	>{\adjustbox{angle=#1,lap=\width-(#2)}\bgroup}%
	l%
	<{\egroup}%
}
\newcommand*\rotangle[2]{\multicolumn{1}{R{#1}{#2}}}

\newcommand{\revised}[1]{#1}
\newcommand{\new}[1]{#1}

\usepackage[capitalize]{cleveref}
\crefname{section}{Sec.}{Secs.}
\Crefname{section}{Section}{Sections}
\Crefname{table}{Table}{Tables}
\crefname{table}{Tab.}{Tabs.}

\def\cvprPaperID{3417} %
\def\confName{CVPR}
\def\confYear{2022}

\begin{document}

\title{SNUG: Self-Supervised Neural Dynamic Garments}
\author{Igor Santesteban$^1$~~~~~~~~~~~Miguel A. Otaduy$^1$~~~~~~~~~~~Dan Casas$^1$\\[0.3cm]
$^1$Universidad Rey Juan Carlos, Spain\\[0.1cm]
{\tt\small first.last@\{urjc.es\}}\\ {\small \url{http://mslab.es/projects/SNUG}}
}

\twocolumn[{%
	\renewcommand\twocolumn[1][]{#1}%
	\maketitle
	\begin{center}
		\vspace{-0.45cm}
		\includegraphics[width=1.0\linewidth]{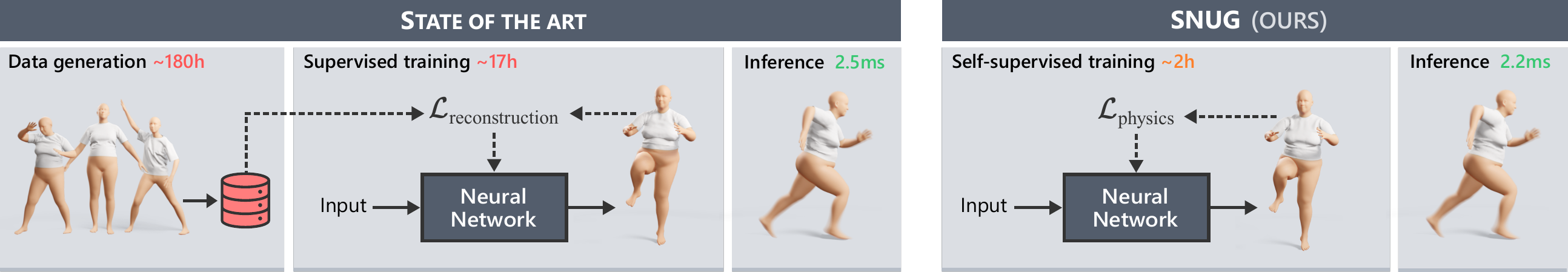}
		\vspace{-0.355cm}
		\captionof{figure}{
			Existing learning-based methods for garment deformations (left) use supervised training schemes that require the expensive computation of large datasets. In contrast, our approach SNUG (right) is a learning-based method that enables the self-supervised training of dynamic neural 3D garments, without requiring any ground-truth data.}
	\label{fig:teaser}
	\end{center}
}]

\thispagestyle{empty}

\begin{abstract}
We present a self-supervised method to learn dynamic 3D deformations of garments worn by parametric human bodies.
State-of-the-art data-driven approaches to model 3D garment deformations are trained using supervised strategies that require large datasets, usually obtained by expensive physics-based simulation methods or professional multi-camera capture setups.
In contrast, we propose a new training scheme that removes the need for ground-truth samples, enabling self-supervised training of dynamic 3D garment deformations.  
Our key contribution is to realize that  physics-based deformation models, traditionally solved in a frame-by-frame basis by implicit integrators, can be recasted as an optimization problem.
We leverage such optimization-based scheme to formulate a set of physics-based loss terms that can be used to train neural networks without precomputing ground-truth data. 
This allows us to learn models for interactive garments, including dynamic deformations and fine wrinkles, with a two orders of magnitude speed up in training time compared to state-of-the-art supervised methods.
\end{abstract}

\vspace{-0.5cm}
\section{Introduction}
The efficient modeling of digital garments is an active area of research due to the large number of applications, including fashion design, e-commerce, virtual try-on, and video games.
The traditional approach to this problem is through physics-based simulation \cite{nealen2006physically}, but the high computational cost required at run time hinders the deployment of these techniques to real-world applications.
Recently, learning-based methods \cite{santesteban2019virtualtryon,patel2020tailor,gundogdu2019garnet,ma2020dressing3d,vidaurre2020fcgnn,tiwari20sizer,wang2018multimodalspace,shen2020garmentgeneration} have demonstrated that it is possible to closely approximate the accuracy of physics-based solutions.
These methods use \textit{supervised} learning strategies to find a function that outputs a deformed garment given an input body descriptor.
During the training phase, the supervision is enforced by directly minimizing at a vertex level the difference between the predicted garment and ground-truth 3D meshes. 
Despite requiring hours of training, learning-based methods are highly-efficient to evaluate at run time, therefore they potentially offer an attractive alternative to traditional physics-based solutions.

However, the need for large datasets in current supervised methods is far from ideal. 
Ground-truth meshes must be obtained --for each combination of garment, body shape, and pose--  via computationally-expensive simulations \cite{narain2012arcsim} or complex 3D scanning setups\cite{pons2017clothcap}, which heavily \new{hinders} the scalability of current learning-based methods.
We observe that for similar image-based problems, \textit{self-supervised} strategies have shown that it is possible to learn complex tasks without requiring ground-truth data \cite{raj2018swapnet,wu2019mm}. Unfortunately, self-supervision for dynamic 3D clothing has not been explored.

In this work, we present a self-supervised method to learn dynamic deformations of 3D garments worn by parametric human bodies.
The key to our success is realizing that the solution to the equations of motion used in current physics-based methods can also be formulated as an optimization problem \cite{Martin2011}.
More specifically, we show that the per-time-step numerical integration scheme used to update the vertex position (\textit{e.g.,} backward Euler) in physics-based simulators, can be recast as an optimization problem, and demonstrate that the function for this minimization can become the central ingredient of a self-supervised learning scheme.
Since this objective function includes both an inertial term and static term directly derived from the equations of motion, we are able to learn time-dependent and pose-dependent deformations \textit{without} any ground-truth data.

The advantages of self-supervision go beyond removing the need for ground-truth data. By reformulating the learning tasks in terms of physics-based intrinsic properties instead of explicit 3D surface similarity, we also mitigate the smoothing artifacts common in supervised methods where L2 losses are used directly at the vertex level \cite{patel2020tailor}.
Additionally, self-supervised approaches also generalize better to test sequences outside the distribution of the training set.
Finally, we also show how different material models can be easily formulated in our self-supervised framework, bringing the generalization capabilities of physics-based solutions (\textit{i.e.}, deform any material) to learning-based methods, without requiring any precalculation or offline step.   

All in all, our main contribution is a novel learning-based method capable of learning to dynamically deform garments using a self-supervised strategy.
We demonstrate the superiority of our approach in terms of data requirements, training time, and inference time, and we quantitatively and qualitatively compare our results with state-of-the-art supervised methods.

\section{Related Work}
Existing methods that model how cloth and garment deform can be categorized into two groups: physics-based models and learning-based models.
\vspace{-0.4cm}
\paragraph{Physics-Based Methods.}
Physics-based simulation of cloth is to date a very mature field. Over the years, many methods have been developed to solve the most relevant challenges.
These include the design of deformation models such as in-plane and bending energies~\cite{Kim2020,Grinspun2003}, robust implicit solvers~\cite{Baraff1998}, rich and efficient contact handling~\cite{Bridson2002,tang2018gpu}, or adaptive discretizations~\cite{narain2012arcsim}.
Recent efforts also include the design of differentiable physics simulators~ \cite{hu2019difftaichi}, including specific problems of cloth simulation such as continuous collision detection and constraint-based solvers~\cite{liang2019differentiable}, \new{and physics-based objectives for tracking and reconstruction of garments
\cite{yu2019simulcap,li3DVphysicsaware}}.
While the majority of the cloth simulation models represent the fabric as a continuum, a recent line of research uses yarn-level representations for high-resolution detail~\cite{Kaldor2008,cirio2014yarn}. Some works also show two-way coupling between garments and soft-body avatars~\cite{romero2020skinmechanics,Montes2020}. While we do not tackle this level of detail in our paper, more accurate methods could be used to replace our cloth and body models.

\vspace{-0.4cm}
\paragraph{Learning-Based Methods.}
In contrast to physics-based models, which typically require solving large systems of nonlinear equations at each time step, learning-based methods aim at estimating a single function that directly outputs the desired deformation for any input.
Inspired by early works on Pose Space Deformation \cite{lewis2000pose}, a common strategy is to learn parametric garment deformations, which are added to a mesh template, as a function of pose \cite{guan2012drape,wang2019intrisicspace}, shape \cite{vidaurre2020fcgnn}, pose-and-shape \cite{santesteban2019virtualtryon,bertiche2020cloth3d}, design \cite{patel2020tailor,wang2018multimodalspace,ma2020dressing3d}, or garment size \cite{tiwari20sizer}.

To this end, state-of-the-art methods for garments use \textit{supervised} strategies that require large datasets of ground-truth data of the specific task to be learned. 
This methodology has been recently explored for many use cases, including 3D reconstruction \cite{alldieck19cvpr,alldieck2018detailed,saito2019pifu,zhu2020deep}, garment design \cite{shen2020garmentgeneration,vidaurre2020fcgnn,wang2018multimodalspace}, animation \cite{bertiche2021ICCV,huang2020arch,wang2019intrisicspace,patel2020tailor,gundogdu2019garnet,ma2020dressing3d}, and virtual try-on \cite{Zhao2021M3DVTONAM,bhatnagar2019multi,santesteban2019virtualtryon,guan2012drape}.
To efficiently tackle the learning task, and depending on the goal of each method, different supervision terms and domains have been used.
Most methods use direct 3D supervision at the vertex level \cite{santesteban2019virtualtryon,patel2020tailor,vidaurre2020fcgnn,gundogdu2019garnet}, but image-based 2D supervision in form of UV maps \cite{lahner2018deepwrinkles,shen2020garmentgeneration,jin2020pixel}, point clouds \cite{Saito:CVPR:2021,ma2021power}, or sketches \cite{wang2018multimodalspace} also exist.
\new{Very recently, implicit representations have shown impressive results on learning to deform humans \cite{deng2020nasa,leapCVPR21,alldieck2021imghum} and dress avatars \cite{Saito:CVPR:2021,tiwari21neuralgif,corona2021smplicit,MetaAvatar:NeurIPS:2021}.}

Datasets are a fundamental piece to enable supervision, and most methods \cite{santesteban2019virtualtryon,patel2020tailor,wang2018multimodalspace,bertiche2020cloth3d} opt for synthetic data generated with physics-based simulators such as ARCSim \cite{narain2012arcsim} or Argus \cite{li2018implicit}. Alternatively, other methods~\cite{lahner2018deepwrinkles,tiwari20sizer,ma2020dressing3d,Saito:CVPR:2021} use high-quality 3D scans obtained in expensive multi-camera setups \cite{zhang2017detailed,pons2017clothcap}. Despite the success of all these supervised methods for learning-based garments, relaying on ground-truth data to train the models is a major limitation due to the associated costs and \new{hinders} to create datasets.

Self-supervised strategies are the ideal alternative to circumvent the need for ground-truth data in learning-based methods \cite{stewart2017label}. 
Instead of relying on losses that evaluate prediction error based on the difference with respect to ground-truth samples, \textit{self-supervised} methods use implicit properties of the training data (or domain) as a supervision signal \cite{zhu2019physics}.
This strategy is nowadays very popular in data-driven methods for image-based problems \cite{zhu2017unpaired,umr2020,raj2018swapnet},
however, almost all state-of-the-art approaches to learn 3D garment deformations rely on ground-truth data \cite{patel2020tailor,santesteban2019virtualtryon,gundogdu2019garnet}. 
For 3D deformations tasks not related to garments, many works use physics laws or constraints as a supervision signal \cite{zhu2019physics,tompson2017accelerating,xie2018tempoGAN}. For example, Tompson \etal \cite{tompson2017accelerating} enforce incompressibility constraints to learn to solve the system of equations required in physics-based fluid simulation, Xie \etal \cite{xie2018tempoGAN} enforce temporal
coherence of consecutive frames in fluid simulations to enhance detail, 
and Zhu \etal \cite{zhu2019physics} incorporates the governing equations of the physical model (\textit{i.e.}, Partial Differential Equations, PDEs) in the loss to learn image-based flow simulations.

Despite the significant progress in self-supervised learning, no previous works addresses the learning of 3D garments in self-supervised strategy, \revised{with just the notable and very recent exception of PBNS \cite{bertiche2021pbns}.}
PBNS proposes to learn pose space deformations for garments by enforcing \textit{static} physical consistency during the training of the model. 
We follow a similar underlying idea, but propose to use a full physics-based deformation scheme recast as an optimization problem to learn, for first time, a model for \textit{dynamic} garment deformations with self-supervision only.
Additionally, our approach learns shape-dependent effects and is able to cope with a material model that produces highly-realistic and finer wrinkles.

\section{Method}
\begin{figure*}
	\centering
	\includegraphics[width=\linewidth]{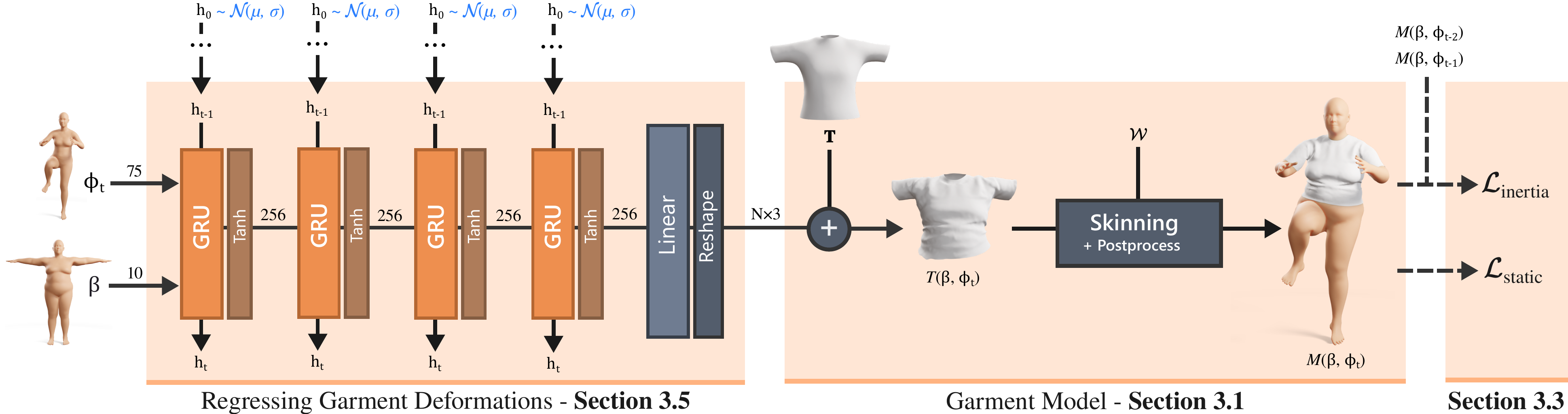}
	\caption{Overview of our method. First, the recurrent regressor predicts per-vertex offsets as a function of body shape and motion. These offsets are added to the garment template which is then skinned to produce the final result. 
	We train the network by optimizing a set of physical properties of the predicted garments, removing need for ground-truth data.
	}
	\label{fig:overview}
\end{figure*}

Our goal is to find a function $M()$ that deforms a 3D garment given the underlying body parameters and motion.
To this end, in \cref{sec:garment-model}, we first describe our garment model used to implement $M()$, which is based on per-vertex dynamic 3D displacements that are added to a rigged template mesh.
Then, in \cref{sec:optim-dynamics}, we direct our attention to an optimization-based formulation of dynamic deformations.
Based on this formulation, in \cref{sec:self-supervision}, we introduce our main contribution and describe a physics-based deformation model that allows us to train a regressor $R()$ for  3D garment displacements.
Importantly, our loss is driven by fundamental physical properties of deformable objects, not by the reconstruction of ground-truth garments, and therefore it enables \textit{self-supervised} learning.
In \cref{sec:material-model} we specify the material model used in the different terms of our loss, and define the relevant energies such as the strain, and bending energies. 
Finally, in \cref{sec:regressor} we describe the recurrent architecture used to implement the regressor $R()$. 
See Figure \ref{fig:overview} for an overview of our method.
\subsection{Garment Model}
\label{sec:garment-model}
Similar to state-of-the-art methods for data-driven garments \cite{santesteban2019virtualtryon,gundogdu2019garnet,patel2020tailor,bertiche2020cloth3d,vidaurre2020fcgnn}, we leverage and extend existing human body models \cite{loper2015smpl,feng2015avatar} to encode garment deformations.
More specifically, we build our representation on top of the popular SMPL human model \cite{loper2015smpl}. SMPL encodes bodies by deforming a rigged human template according to shape and pose-dependent deformations that are learned from data. Following this idea, we define our garment model as
\begin{align}
M(\upbeta, \upphi) &= W(T(\upbeta, \upphi), J(\upbeta), \uptheta,\mathbf{W}_{\text{G}}) \\
T(\upbeta, \upphi) & = \mathbf{T} + R(\upbeta, \upphi) \label{eq:model_in_rest}
\end{align}
where $W$ is a skinning function (\textit{e.g.}, linear blend skinning or dual quaternion) with skinning weights 
$\mathbf{W}_{\text{G}}$, joint locations $J(\upbeta)$, and motion parameters $\upphi$ that articulate an unposed deformed garment mesh $T(\upbeta,\upphi)$.  
The latter is computed from a garment template mesh $\mathbf{T}$ deformed by a function $R(\upbeta, \upphi)$ that outputs per-vertex 3D displacements to encode dynamic deformations conditioned to the underlying body shape $\upbeta$ and body motion $\upphi$. \revised{The body motion $\upphi$ contains the current body pose $\uptheta$ as well as the global velocity of the root joint.}

Assuming that the garment template $\mathbf{T}$ is correctly located on top of the mean SMPL body mesh \cite{loper2015smpl}, we define  $\mathbf{W}_{\text{G}}$ by borrowing the SMPL skinning weights of closest body vertex in rest pose. 
In the remainder of this section we introduce our novel strategy to learn the 3D displacement regressor $R(\upbeta, \upphi)$.

\subsection{Optimization-Based Dynamic Deformation}
\label{sec:optim-dynamics}
Our goal is to learn the 3D displacement regressor $R(\upbeta, \upphi)$ in Equation \ref{eq:model_in_rest} using a self-supervised strategy. To this end, our first task is to find a set of physics-based properties that describe how cloth behaves. 
Physics-based simulators traditionally solve dynamics by applying a numerical integration scheme, \textit{e.g.}, backward Euler, to the differential equations of motion, and finding the roots of the resulting nonlinear discrete  equations~\cite{nealen2006physically}. This formulation is applied independently at each simulation frame, to iteratively update the positions and velocities of garment vertices.
Our key observation is to realize that the solution to the equations of motion discretized with backward Euler can also be formulated as an optimization problem \cite{Martin2011}, and the objective function for this minimization can become the central ingredient of a self-supervised learning scheme. 
Optimization-based dynamics have been used in the Computer Graphics literature to increase the efficiency and robustness of dynamics solvers, through quasi-Newton schemes and step-size selection \cite{gast2015optimization,Liu2017}. Instead, we propose to leverage such optimization-based formulation to define a loss for training a neural network that generalizes well to any input (\textit{i.e.,} any body shape and motion).

The equations of motion can be discretized with backward Euler as
\begin{align}
\textbf{M}\frac{\mathbf{x}^{t+1} - \mathbf{x}^t - \Delta t \mathbf{v}^t}{\Delta t^2} = \mathbf{f}\left(\mathbf{x}^{t+1} ,\frac{\mathbf{x}^{t+1} - \mathbf{x}^{t}}{\Delta t} \right),
\label{eq:fma}
\end{align}
where $\textbf{M}$ is the mass matrix, $\mathbf{f}$ are forces, and $\mathbf{x}$ and $\mathbf{v}$ are the positions and velocities of garment nodes. The solution to these equations can be recast as an optimization \cite{Martin2011,gast2015optimization}:
\begin{align}
	 \mathbf{x}^{t+1} = \underset{\mathbf{x}}{\arg\min}~~ \frac{1}{2\Delta t^2}(\mathbf{x} - \hat{\mathbf{x}})^\top\textbf{M}(\mathbf{x} - \hat{\mathbf{x}}) + \mathbf{\Phi},
	 \label{eq:minimization}
\end{align} 
where $\hat{\mathbf{x}} = \mathbf{x}^{t} + \Delta t\mathbf{v}^{t}$ is a tentative (explicit) position update, and $\mathbf{\Phi}$ is the potential energy due to internal and external forces $\mathbf{f}$ of the system. %

\subsection{Turning Dynamics into Self-Supervision}
\label{sec:self-supervision}
The key to our method is to define a set of losses based on Equation \ref{eq:minimization} to train the regressor $R()$. 
To this end, we propose a loss with two terms
\begin{align}
	\mathcal{L}=\mathcal{L}_\text{inertia} + \mathcal{L}_\text{static},
	\label{eq:loss} 
\end{align}
where $\mathcal{L}_\text{inertia}$ models the inertia of the garment and it is defined analogous to the first term of Equation \ref{eq:minimization}
\begin{align}
	\mathcal{L}_\text{inertia} = \frac{1}{2\Delta t^2}(\mathbf{x} - \hat{\mathbf{x}})^\top\textbf{M}(\mathbf{x} - \hat{\mathbf{x}}).
	\label{eq:inertia_loss}
\end{align}
Intuitively, this term prevents the change of garment velocities over time, but garment velocities will change anyway due to the underlying body motion, which makes dynamics and wrinkle effects appear.

$\mathcal{L}_\text{static}$, the second term of our loss $\mathcal{L}$, models the potential energy $\mathbf{\Phi}$ of Equation \ref{eq:minimization} which represents the internal and external forces that affect the garment.
Inspired by works from cloth simulation literature \cite{narain2012arcsim,sifakis2012fem}, we define $\mathcal{L}_\text{static}$ as the sum of different physics-based terms that model the energies that emerge on deformable solids, including strain, bending, gravity, and collisions
\begin{align}
	\mathcal{L}_\text{static} = \mathcal{L}_\text{strain} + \mathcal{L}_\text{bending} + \mathcal{L}_\text{gravity} + \mathcal{L}_\text{collision}.
	\label{eq:static_loss}
\end{align}
This formulation of $\mathcal{L}_\text{static}$ is general, and the definition of each term depends on the material model used, which we detail in the next section.
\subsection{Material Model}
\label{sec:material-model}
The literature of simulation of elastic solids characterizes materials using equations that relate stimuli (\textit{e.g.}, deformations) to material response (\textit{e.g.}, energies) \cite{sifakis2012fem}. 
Inspired by this, and with the goal of learning physically-correct garment behaviors, we define the terms of our static loss $\mathcal{L}_{\text{static}}$ based on equations of state-of-the-art cloth simulators \cite{narain2012arcsim} to model the following energies:
\paragraph{Membrane Strain Energy.} The membrane strain term models the response of the material to in-plane deformation. Given a deformed position $x \in \mathbb{R}^3$ and an undeformed position $X \in \mathbb{R}^2$ (\textit{i.e.}, the garment template), it defines an internal energy based on a first-order deformation metric, typically the deformation gradient $\mathbf{F} =  \frac{\partial x}{\partial X} $. 
In our loss we implement it using the Saint Venant Kirchhoff (StVK) elastic material model that defines membrane strain energy as
\begin{align}
	\Psi_\text{S} = \frac{\lambda}{2}\text{tr}(\mathbf{G})^2 + \mu\text{tr}(\mathbf{G}^2),
	\label{eq:strain}
\end{align}
 where $\lambda$ and $\mu$ are the Lam\'{e} constants, and $\mathbf{G} = \frac{1}{2}(\mathbf{F}^\top\mathbf{F} - \mathbf{I})$ is the Green strain tensor. %
 The membrane strain energy of the mesh is computed as
 \begin{align}
\mathcal{L}_\text{strain}=\sum_\text{triangles} \mathbf{V}\Psi_\text{S},
\end{align}
where $\mathbf{V}$ is the volume of each triangle (\textit{i.e.,} area $\times$ thickness).
\paragraph{Bending Energy.} The bending term models the energy due to the angle of two adjacent faces and we model it as  
\begin{align}
	\mathcal{L}_\text{bending} = \sum_\text{edges} \frac{k_\text{bending}}{2}\theta^2
\end{align}
where $\theta$ is the dihedral angle between the faces and $k_\text{bending}$ is a bending stiffness.
\paragraph{Gravity.} To model the effect of gravity in the learned deformations, we add a loss term with the potential energy of each cloth vertex 
\begin{align}
	\mathcal{L}_\text{gravity} = \sum_\text{vertices} - m \, \mathbf{g}^\top x
\end{align}  
where $m$ is the vertex mass, and $\mathbf{g}$ is the  gravitational acceleration.

\paragraph{Collision Penalty.}  This term is crucial to learn plausible deformations, enforcing the garment to follow the underlying body motion. We implement it as
\begin{align}
	\mathcal{L}_\text{collision} = \sum_\text{vertices} k_\text{collision}~ \text{max}(\upepsilon - d(x), 0)^3
\end{align}
where $d(x)$ is a function that computes the distance to the body, $k_\text{collision}$ is a collision stiffness, and $\upepsilon$ is a safety margin to prevent the garment from overlapping with the body surface.

\begin{figure}
 \begin{subfigure}{0.49\linewidth}
    \includegraphics[width=\linewidth]{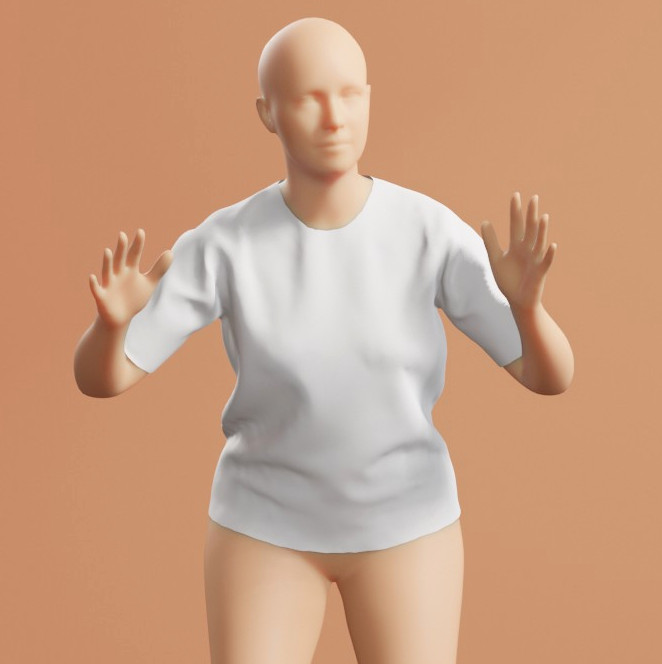}
    \caption{StVK (Ours)} \label{fig:1a}
  \end{subfigure}%
  \hspace*{\fill}   %
  \begin{subfigure}{0.49\linewidth}
    \includegraphics[width=\linewidth]{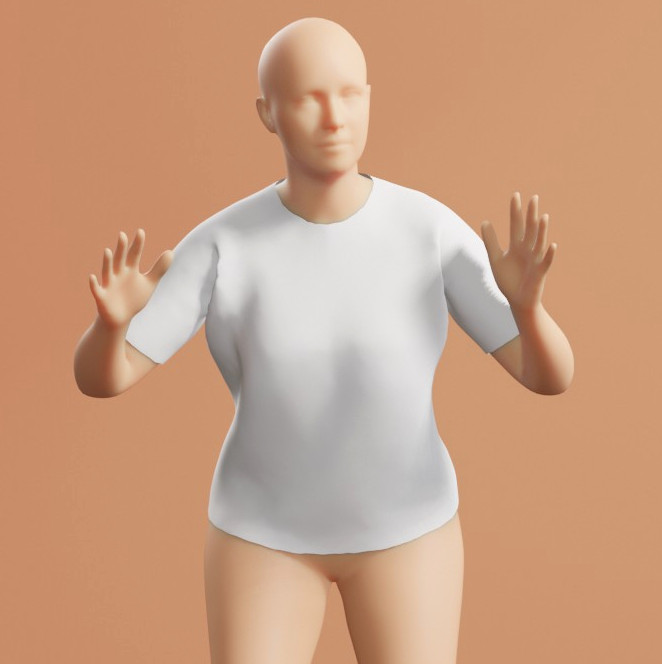}
    \caption{Mass-spring (PBNS \cite{bertiche2021pbns})} \label{fig:1b}
  \end{subfigure}%
	\caption{The material model used is crucial to obtain realistic garment behaviors. We formulate our losses using the Saint Venant Kirchoff (StVK) model, in contrast to simpler alternatives that lead to less expressive deformations.}
	\label{fig:material_model}
\end{figure}

\vspace{0.3cm}
To highlight the realism of the proposed material, in Figure \ref{fig:material_model} we show a ground-truth simulation of our model, and the simpler material model used in PBNS\cite{bertiche2021pbns} based on a traditional mass-spring formulation.
Overall, our model is capable of reproducing more complex behaviors typically present in garments, including wrinkles and folds at different scales.

\subsection{Regressing Garment Deformations}
\label{sec:regressor}
With our novel self-supervised loss $\mathcal{L}$ defined in Section \ref{sec:self-supervision}, we are ready to train the garment displacement regressor $R()$ from Equation \ref{eq:model_in_rest} without requiring ground-truth data.
To this end, in order to model the time dependencies of the inertial term $\mathcal{L}_{\text{inertia}}$, we implement the regressor using 4 Gated Recurrent Units (GRU), each with an output of size 256, and \texttt{tanh} as the activation function (see Figure \ref{fig:overview}). 
However, the recurrent nature of GRUs combined with the lack of ground-truth values to guide the training process make the regressor converge to bad solutions if a naive recurrent training protocol is used. We need to take special care into how the hidden states of the GRUs are initialized and updated. 

Intuitively,
the model should be able to learn dynamics from just 3 frames, since $\mathcal{L}_{\text{inertia}}$ from Equation \ref{eq:inertia_loss} depends only on the vertex positions and velocities of the previous step.
Therefore, we train our network using sub-sequences of 3 frames. Interestingly, we found that training on longer sub-sequences also minimizes $\mathcal{L}_{\text{inertia}}$ correctly, but the learned deformations do not model true dynamics.
  
At runtime, the network supports sequences of arbitrary length, but results can degrade noticeably for sequences longer than those used in training if initialization of the GRU hidden states is not well handled. 
More specifically, we observe that for each training sub-sequence, setting the initial hidden states $\mathbf{h}_0 = 0$ \new{hinders} the network to generalize to sequences longer than 3 frames. 
We address this issue by sampling the initial state $\mathbf{h}_0$ of each GRU from $\mathcal{N}(\mu, \sigma)$ (empirically, $\mu=0$ and	$\sigma=0.1$), which allows the model to generalize well even for sequences with thousands of frames. 
Notice that at runtime the state $\mathbf{h}_t$ depends on an arbitrarily large number of previous frames, not just the last 3, hence the use of noise to initialize states on train sub-sequences is fundamental to augment variance in states.

\section{Evaluation}

\subsection{Training}
\label{sec:training}
To self-supervise the training process of our regressor $R()$ we need to feed it with human motions and shapes.
To this end, we use a set of 52 sequences from the AMASS dataset \cite{AMASS:ICCV:2019}, totaling 6,519 frames, which we split into sub-sequences of 3 frames as described in Section \ref{sec:regressor}. We set aside 4 full sequences for validation purposes.
To provide body shape variety at train time, each of the sub-sequences is assigned a different body shape $\upbeta$ sampled from $\mathcal{U}(-3,3)$ at each epoch.
Notice that, enabled by our self-supervised approach, this strategy allows us to train using thousands of different body shapes, while competitive supervised methods are limited to a dramatically smaller shape sample (\cite{patel2020tailor} uses 9 shapes, \cite{santesteban2019virtualtryon} uses 17) due to the computational restrictions caused by the need for a ground-truth database.

Regarding the network hyper-parameters, we use a batch size of 16, initially train for 10 epochs using a learning rate of 0.001, and then resume the learning with a learning rate of 0.0001 until it converges.
\new{This approach is fast, works for all garments, and avoids erroneous states. The rest of the material and training parameters do not affect stability.
Larger learning rates can introduce instabilities due to energy spikes that make the training struggle to recover (\ie the predicted mesh has collisions that are too large to be resolved).  
Small body-garment collisions are not a problem -- \eg, we can handle pants despite self-collisions in the legs on some poses.}

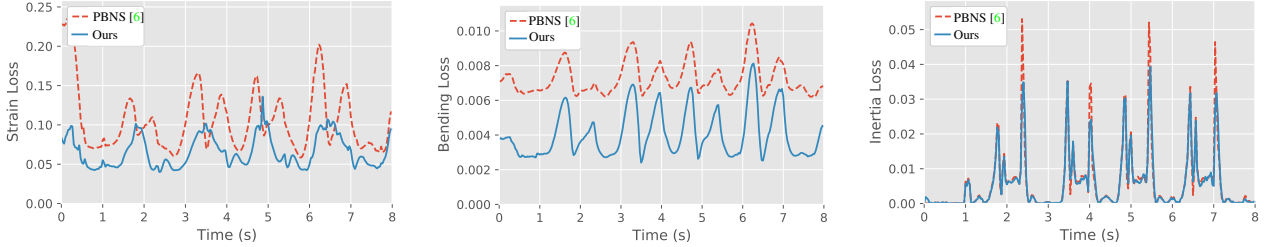
\begin{figure*}
	\centering
	\begin{tikzpicture}
		\node[] (image) at (0,0.1)
		{\includegraphics[width=0.31\textwidth,trim=0pt 0 0 0, clip]{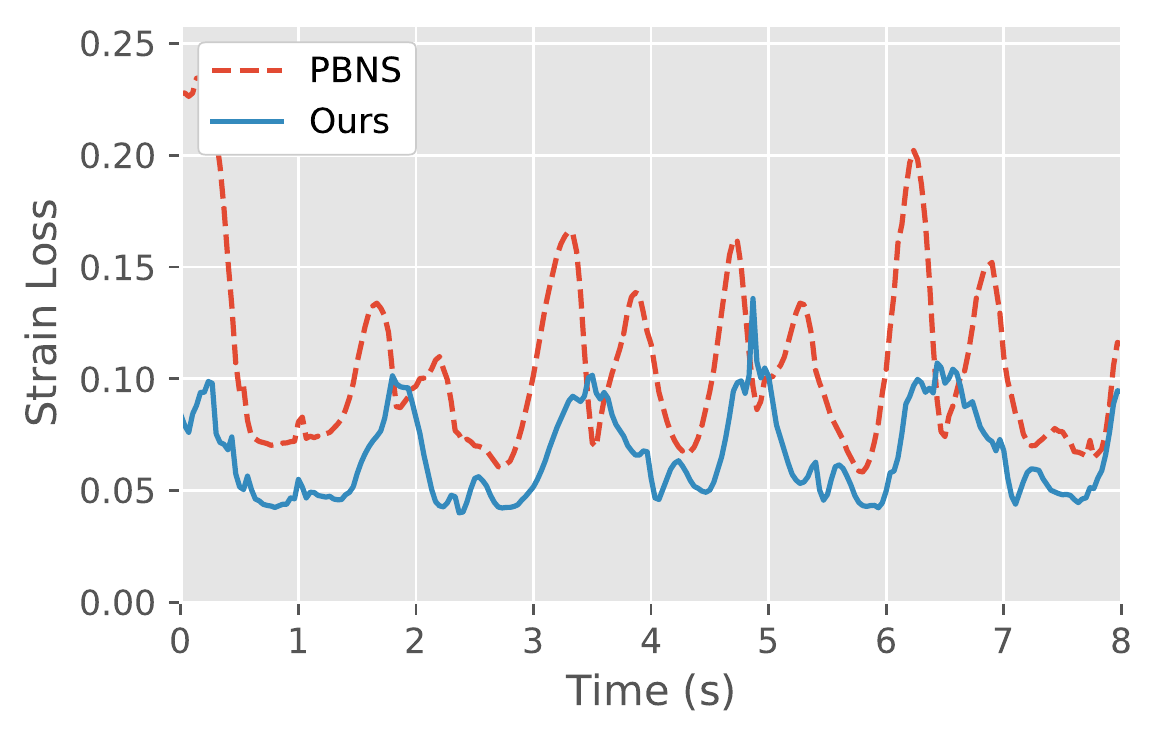}};
		\filldraw[fill=white, draw=white] (-1.5,1.62) rectangle (-0.8,1.12);
		\node[right of=image, xshift=-1.42cm,yshift=0.48cm] (textB) at (-1,) {
		\scalebox{0.55}{	\small{\hspace{3.2em} PBNS \cite{bertiche2021pbns}}}};
	\node[right of=image, xshift=-1.60cm,yshift=0.26cm] (textC) at (-1,) {
		\scalebox{0.55}{	\small{\hspace{3.2em} Ours}}};
	\end{tikzpicture}
	\begin{tikzpicture}
	\node[] (image) at (0,0.1)
	{\includegraphics[width=0.31\textwidth,trim=0pt 0 0 0, clip]{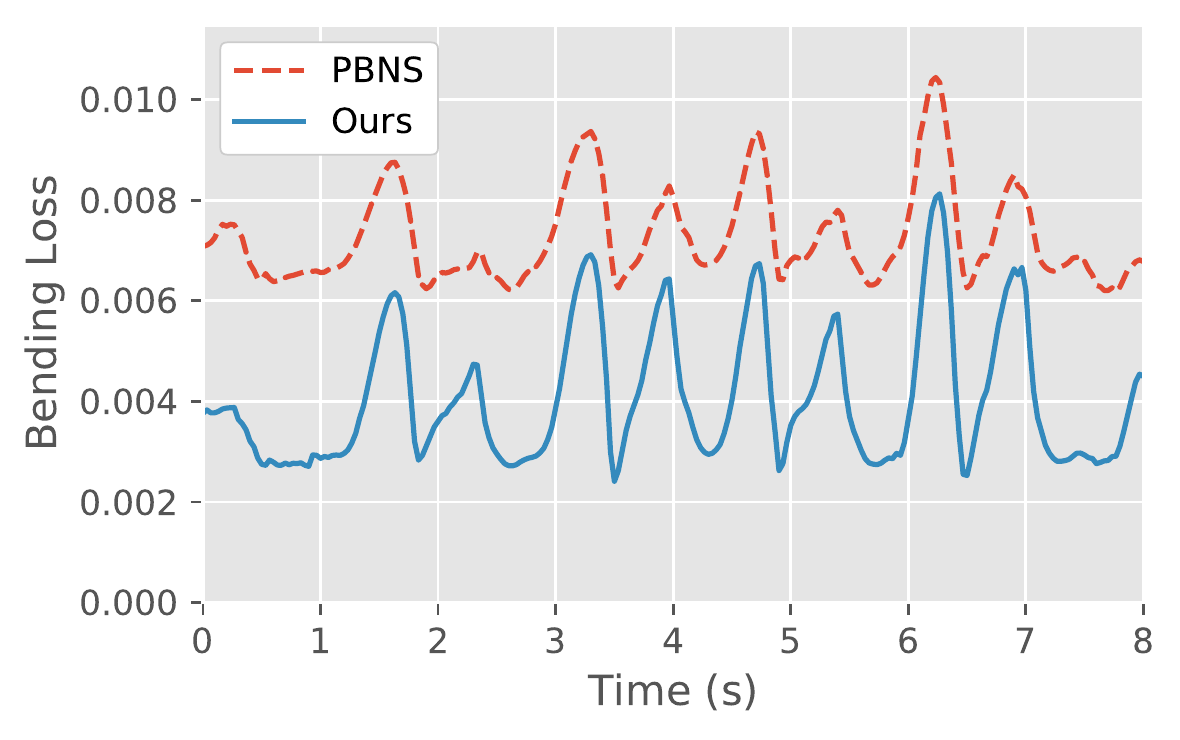}};
	\filldraw[fill=white, draw=white] (-1.4,1.59) rectangle (-0.7,1.12);
	\node[right of=image, xshift=-1.365cm,yshift=0.46cm] (textB) at (-1,) {
		\scalebox{0.55}{	\small{\hspace{3.2em} PBNS \cite{bertiche2021pbns}}}};
	\node[right of=image, xshift=-1.53cm,yshift=0.26cm] (textC) at (-1,) {
		\scalebox{0.55}{	\small{\hspace{3.2em} Ours}}};
\end{tikzpicture}
	\begin{tikzpicture}
	\node[] (image) at (0,0.1)
	{\includegraphics[width=0.31\textwidth,trim=0pt 0 0 0, clip]{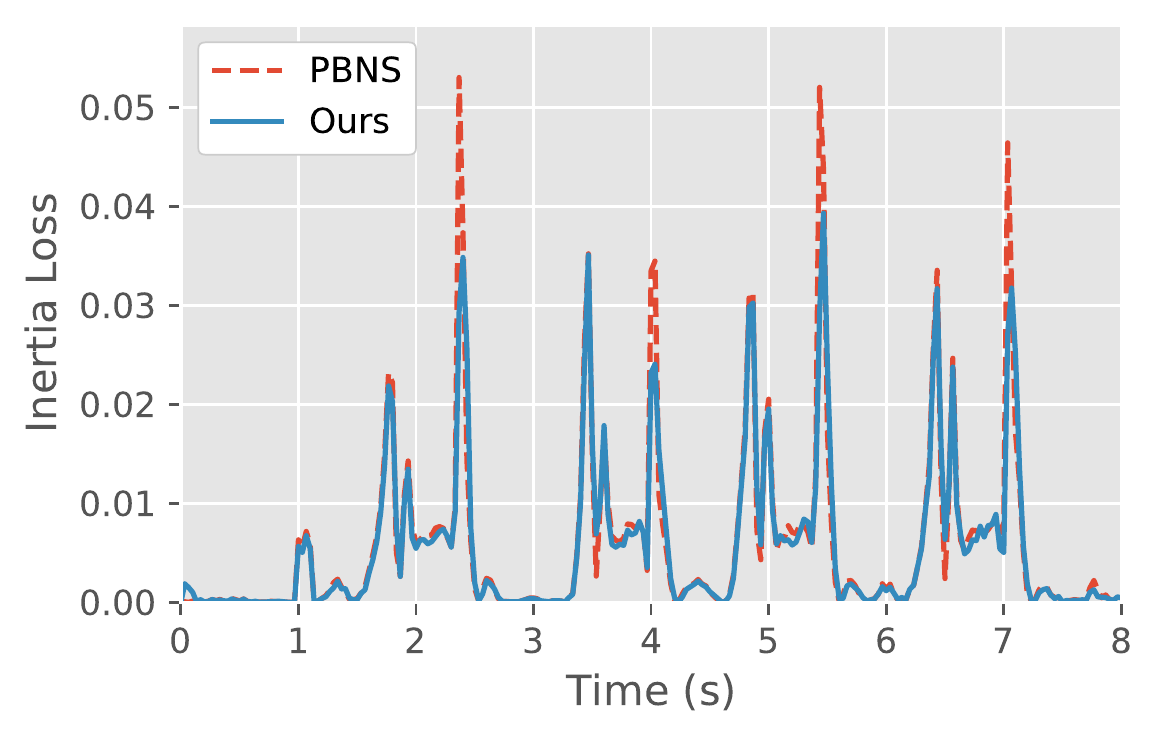}};
	\filldraw[fill=white, draw=white] (-1.5,1.62) rectangle (-0.8,1.154);
	\node[right of=image, xshift=-1.45cm,yshift=0.475cm] (textB) at (-1,) {
		\scalebox{0.55}{	\small{\hspace{3.2em} PBNS \cite{bertiche2021pbns}}}};
	\node[right of=image, xshift=-1.62cm,yshift=0.26cm] (textC) at (-1,) {
		\scalebox{0.55}{	\small{\hspace{3.2em} Ours}}};
\end{tikzpicture}
	\caption{Quantitative evaluation of our approach. We evaluate the minimization error in different the physics-based terms used in our loss, in the test sequence \texttt{01\_01} of AMASS \cite{AMASS:ICCV:2019}. Sudden motion changes (\textit{e.g.}, jumps) naturally produce peaks in the inertial term, due to drastic changes in the velocity of the garment. Intuitively, cloth dynamics arise when the garment resists to those changes enforced by the body, therefore lower inertial values indicates that our model better learns time-dependent effects than PBNS \cite{bertiche2021pbns}. See the supplementary video for qualitative results of this evaluation. The significantly improved realism of our method is better appreciated looking into sequences of deformed garments. 
	}
	\label{fig:quantitive_dynamics}
\end{figure*}

Our approach does not require balancing loss terms, we just need to set the material properties of the garment. 
\new{To this end, we tune material parameters to produce a desired fabric behavior, hence the parameters of the loss have a physical meaning -- they are not arbitrary hyperparameters}. 
To compute the mass matrix $\mathbf{M}$ we use real measurements of the thickness and density of 100\% cotton fabric ($0.47\text{ mm}$ and $426\text{ kg/m}^3$ respectively). 
The rest of the material parameters have the following values: the Lam\'{e} constants are set to $\lambda=20.9$ and $\mu=11.1$, the bending stiffness $k_\text{bending}=3.96\mathrm{e}{-5}$, the collision stiffness $k_\text{collision}=250$, and the collision margin $\upepsilon=2\text{ mm}$. We use the same parameters for all our garments.

\revised{To thoroughly validate our model, in addition to comparisons to SOTA methods, in this section we also include ablations and comparisons that use a ground-truth simulated dataset.
For as fair as possible evaluations, such dataset is created using the same motions, and the same train-test split, that we use to train SNUG.
}

We implement our method in a regular desktop PC equipped with an AMD Ryzen 7 2700 CPU, an Nvidia GTX 1080 Ti GPU, and 32GB of RAM.

\begin{table}
	\centering
	\begin{tabular}{rrccc} 
		\toprule 
		& Strain & Bending & Gravity & Inertia \\
		\midrule
		PBNS \cite{bertiche2021pbns}& 0.111 & 0.007 & 0.044 & 0.0035\\
		SNUG (Ours) & \textbf{0.064} & \textbf{0.004} & \textbf{0.028} & \textbf{0.0034}\\
		\bottomrule
	\end{tabular}
	\caption{To quantitative evaluate our method we compute the physics-based loss terms of our trained model, in unseen sequences, and compare to PBNS. We produce lower errors in all terms, indicating that our approach results in deformations that better match physics-based simulators.}
	\label{table:quantitative_evaluation}
\end{table}
\begin{table}[]
	\setlength{\tabcolsep}{5.5pt}
	\begin{tabular*}{\linewidth}{@{\extracolsep{\fill}}lccccc}
		& 
		\rotangle{25}{0em}{W/o bending}
		&  
		\rotangle{25}{0em}{W/o strain}
		&
		\rotangle{25}{1.5em}{W/o gravity}
		&
		\rotangle{25}{0.5em}{W/o inertia}
		&
		\rotangle{25}{0em}{Full} \\
		\midrule
		Mean-curvature error & 17.3  & 19.1 & 7.5 & 2.8 & \textbf{2.7} \\
		\bottomrule
	\end{tabular*}
	\vspace{-0.2cm}
	\caption{
		\revised{Quantitative ablation study. Each term of our loss contributes to the accuracy of the final result.}
	}
	\label{table:ablation_quantitative}
\end{table}
\begin{table}
	\centering
	{
		\setlength{\tabcolsep}{3.2pt}
		\begin{tabular}{rrrrr} 
			\toprule 
			& \multicolumn{1}{c}{Data} & \multirow{2}{*}{Train} & \multirow{2}{*}{Runtime} & \multirow{2}{*}{Memory} \\[-1.5pt]
			& \multicolumn{1}{c}{generation} & & \\
			\midrule
			TailorNet \cite{patel2020tailor} & 29 h~~ & 6.5 h & 10.1 ms & 2114 MB \\
			Santesteban \cite{santesteban2019virtualtryon}& 180 h~~ & 17 h & 2.5 ms & 109 MB\\
			SNUG (Ours) & \textbf{0 h}~~ & \textbf{2 h} & \textbf{2.2 ms} & \textbf{19 MB}\\
			\bottomrule
		\end{tabular}
		\caption{Timings, memory requirements, and performance of state-of-the-art methods. Our self-supervised approach avoids the expensive cost of data generation, while also achieving significantly lower training times. 
		}
		\label{table:performance}
	}
\end{table}

\subsection{Quantitative Evaluation}
To quantitatively evaluate our approach, we measure 
the physics-based terms of our loss $\mathcal{L}$ in test motions and compare it with the predictions of PBNS \cite{bertiche2021pbns}.
Notice that the original PBNS method uses a different (and simpler) material model but, in order to get a meaningful quantitative comparison, we extended and re-trained the publicly available PBNS implementation with our material model defined in Section \ref{sec:material-model}.
Also, notice that we cannot provide this comparison for supervised state-of-the-art methods (\textit{e.g.}, \cite{patel2020tailor,santesteban2019virtualtryon,gundogdu2019garnet}) because the simulation schemes, material models, and parameters used to build their datasets are different and, therefore, the ground-truth physics properties (\textit{i.e.}, our loss terms) might differ significantly. 

Figure \ref{fig:quantitive_dynamics} shows the quantitative evaluation for the most important terms of our loss, and compares it with the extended implementation of PBNS \cite{bertiche2021pbns} using our material, in the test sequence \texttt{01\_01} of AMASS \cite{AMASS:ICCV:2019}.
Notice how our method consistently produces lower error values across all terms (strain, bending, and inertia), indicating that test samples processed with SNUG better match the behavior of physics-based solutions (\textit{i.e.}, the minimization of the terms).
Table \ref{table:quantitative_evaluation} presents a quantitative evaluation of both methods in our full test set (4 sequences, 598 frames unseen at train time), which further demonstrates that our approach improves upon the method of PBNS. 

\revised{To validate each term of our formulation, in Table \ref{table:ablation_quantitative} we show an ablation of the mean-curvature error, evaluated in the test set of our ground truth simulated dataset, when leaving out some of the terms.}

Finally, in Table \ref{table:performance} we also evaluate the memory requirements, training time, and runtime performance of our approach and compare to existing state-of-the-art supervised methods. 
\new{Even if these methods do no address exactly the same problem (\eg TailorNet \cite{patel2020tailor} models garment variations and SNUG does not, but the latter models dynamics)}, SNUG outperforms supervised methods by a large margin in all metrics, resulting in a compact model, only 19MB, trained in just 2h, which opens the door to scalable learning-based garment models.

\begin{figure}
     \centering
     \begin{subfigure}[t]{0.32\columnwidth}
         \centering
         \includegraphics[trim={25cm, 14cm, 25cm, 9cm},clip=true,width=\textwidth]{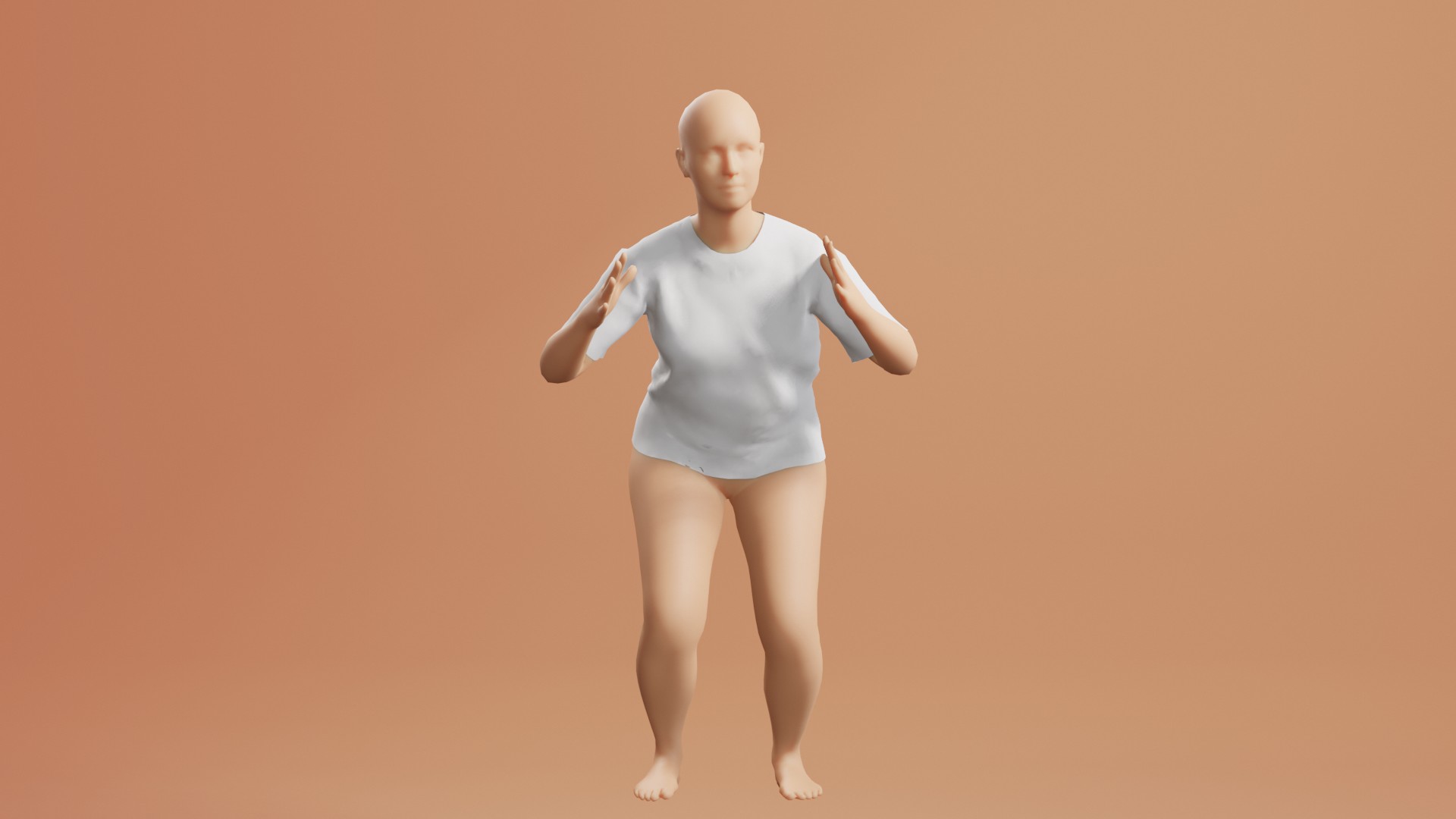}
         \caption{Supervised}
     \end{subfigure}
     \hfill
     \begin{subfigure}[t]{0.32\columnwidth}
         \centering
         \includegraphics[trim={25cm, 14cm, 25cm, 9cm},clip=true,width=\textwidth]{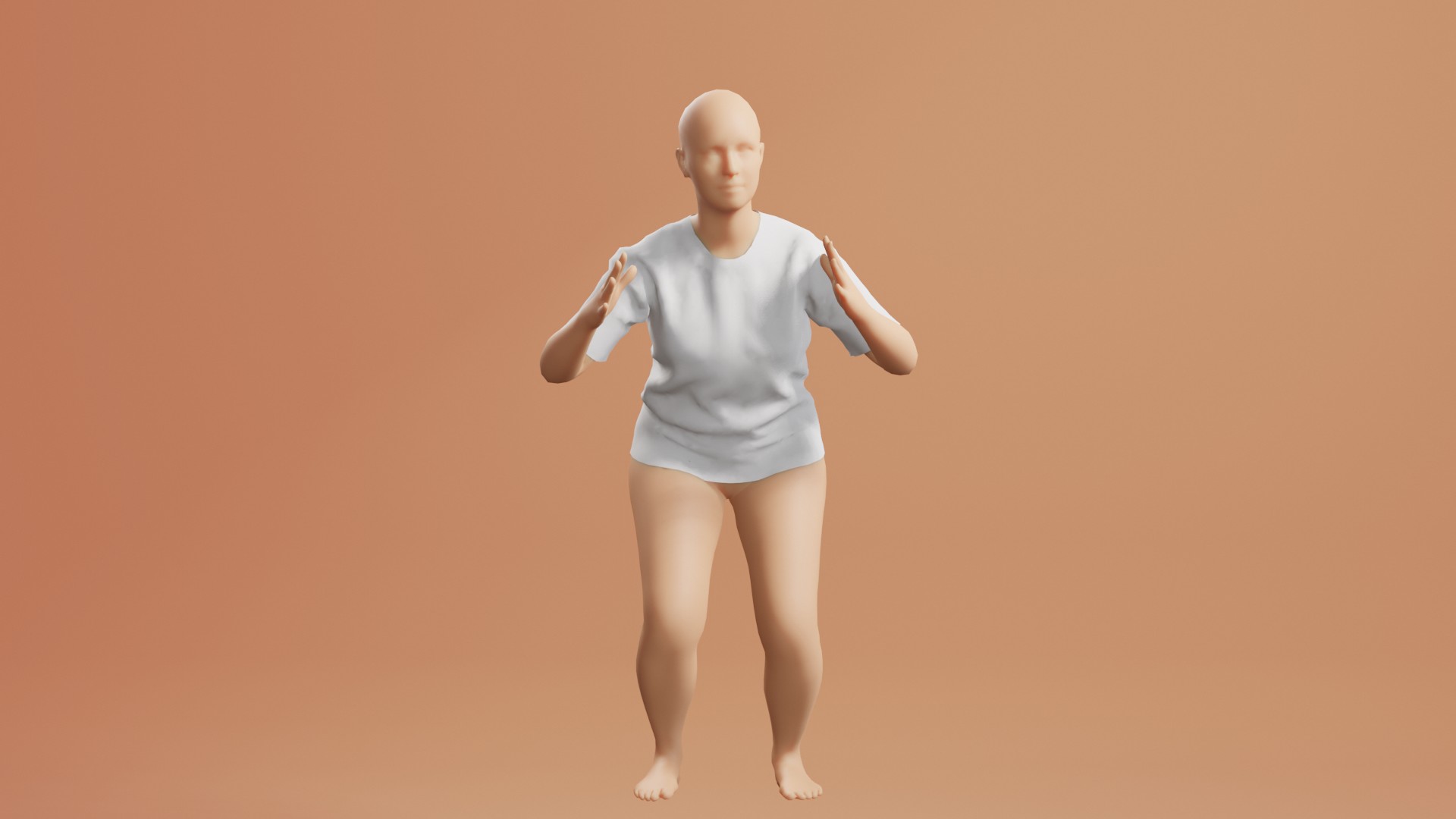}
         \caption{Ours}
     \end{subfigure}
     \hfill
     \begin{subfigure}[t]{0.32\columnwidth}
         \centering
         \includegraphics[trim={25cm, 14cm, 25cm, 9cm},clip=true,width=\textwidth]{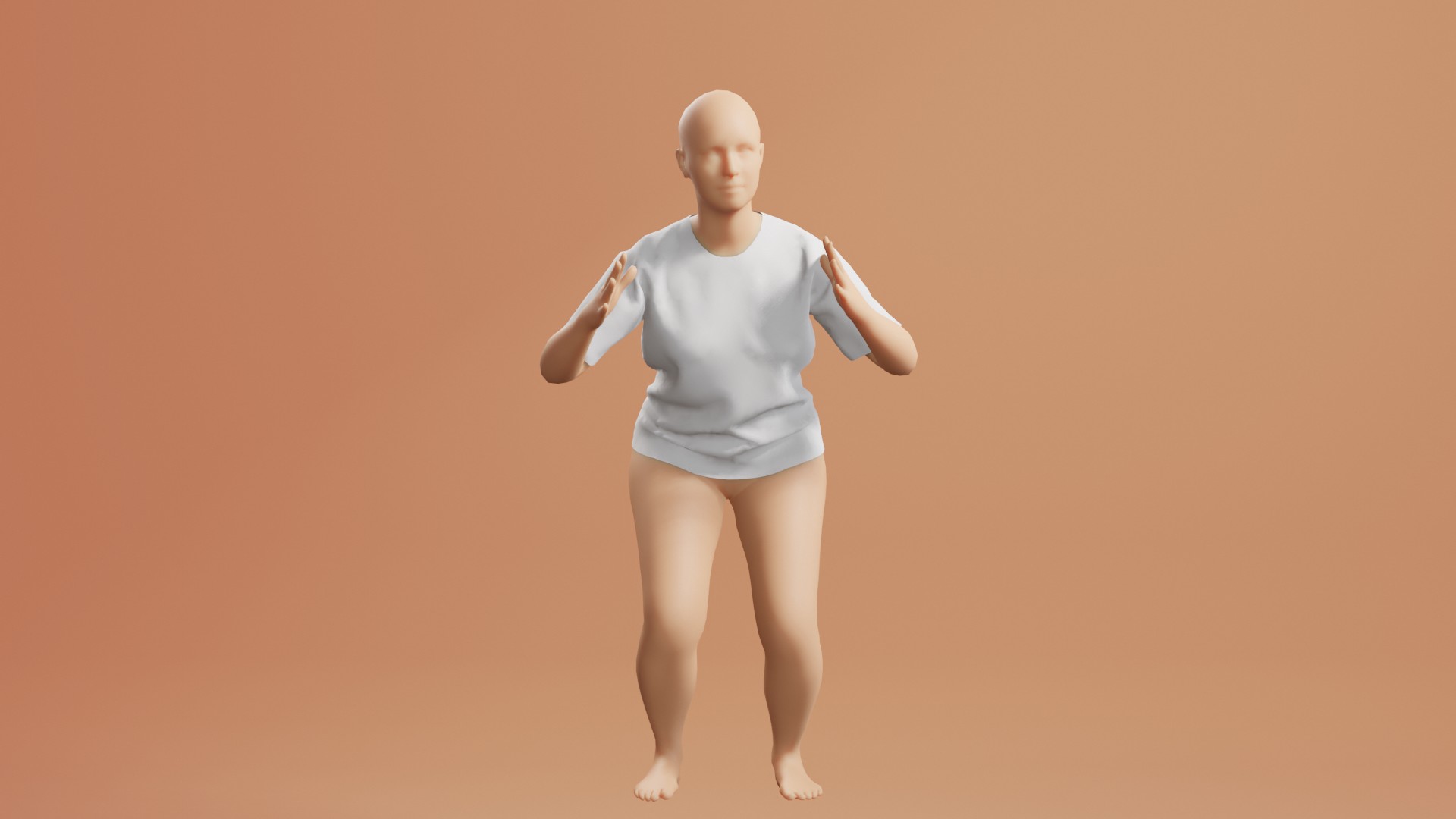}
         \caption{Simulation}
     \end{subfigure}
        \caption{\revised{When trained using same motions and same architecture, direct supervision at the vertex level leads to smoothing artifacts (a). In contrast, our physics-based loss is able to learn more realistic details (b), as shown in this frame from a test sequence.  
        }
        }
        \label{fig:supervised}
\end{figure}

\begin{figure*}[ht]
	\centering
	\includegraphics[width=\linewidth]{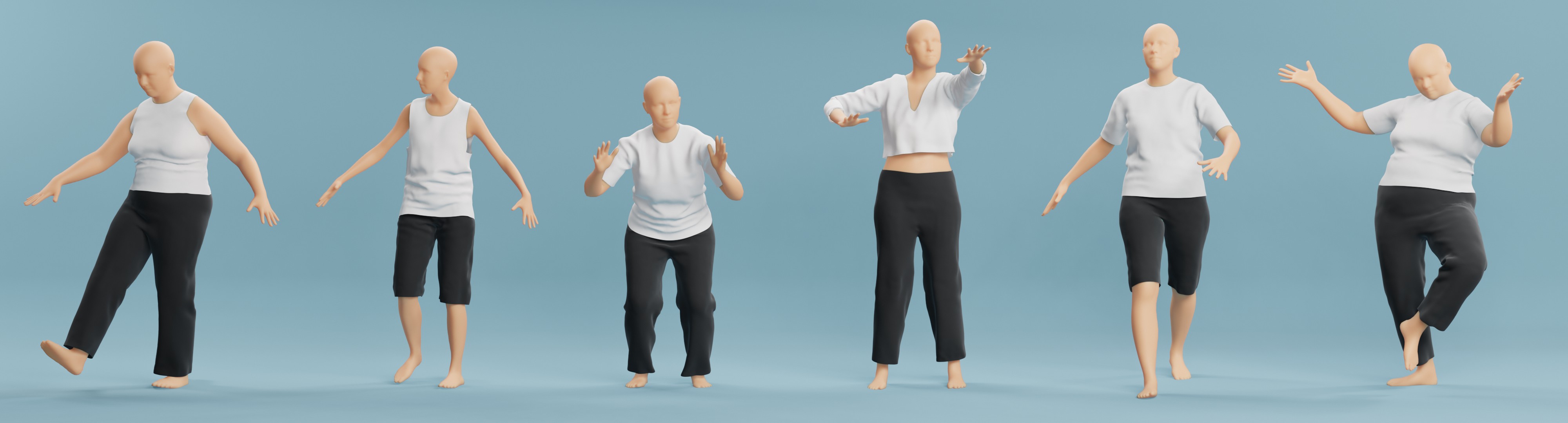}
	\caption{Qualitative results of our self-supervised method, in validation body shapes and poses unseen during training. SNUG successfully learns highly-realistic garment deformations, including fine wrinkles, as a function of body shape and motion.}
	\label{fig:extra_garments}
\end{figure*}
\begin{figure*}[ht]
	\centering
	\vspace{4pt}
	\begin{tikzpicture}
		\node[] (image) at (0, 0)
		{\includegraphics[width=\textwidth]{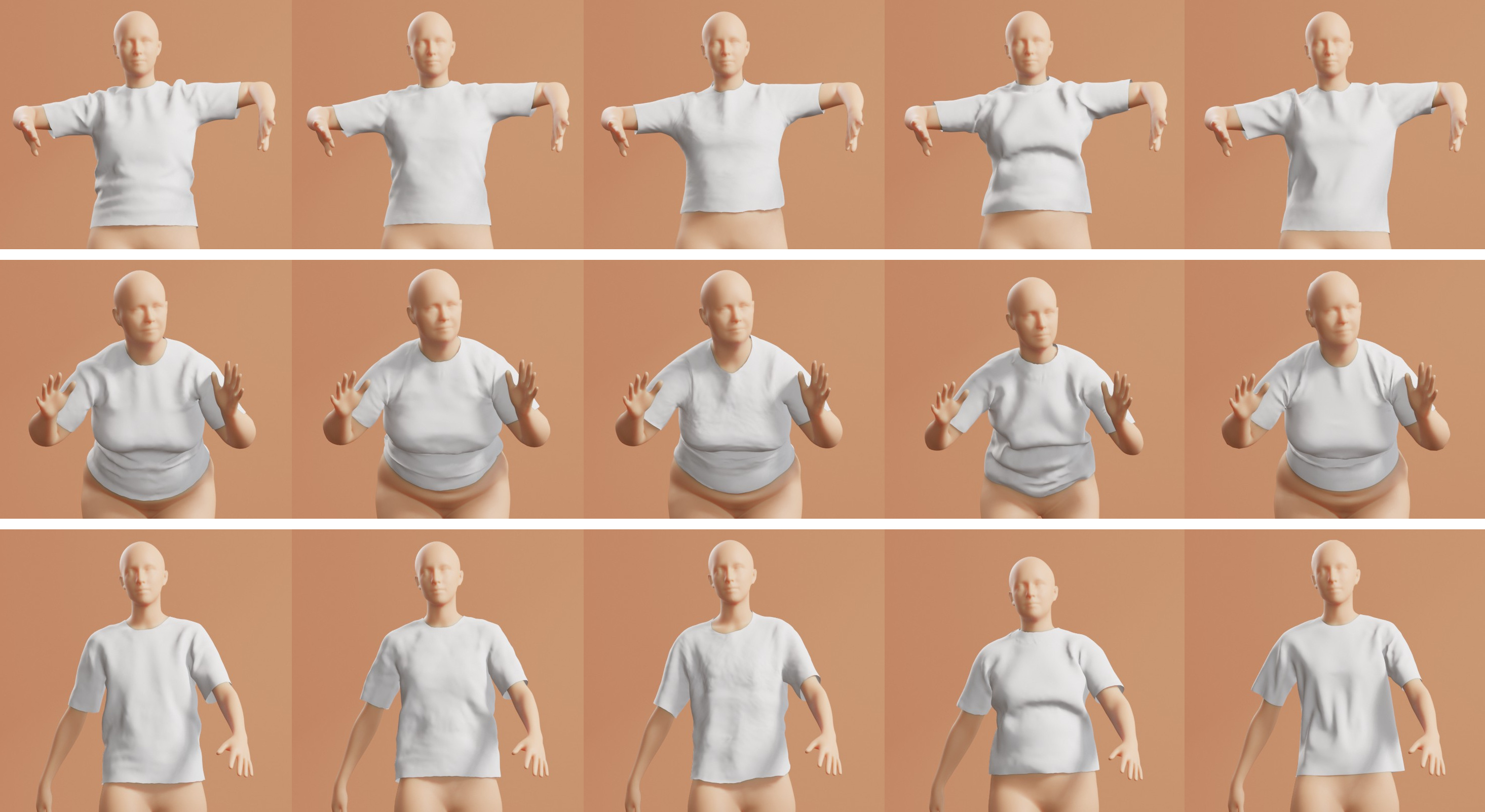}};

		\node[right of=image, xshift=-7.0cm,yshift=5.1cm] (textA) at (-1, 0) {
			\small{SNUG (Ours)}};

		\node[right of=image, xshift=-3.6cm,yshift=5.1cm] (textB) at (-1, 0) {
			\small{Santesteban \textit{et al.} \cite{santesteban2019virtualtryon}}};
		
		\node[right of=image, xshift=0.0cm,yshift=5.1cm] (textC) at (-1, 0) {
			\small{TailorNet \cite{patel2020tailor}}};
		
		\node[right of=image, xshift=3.4cm,yshift=5.1cm] (textD) at (-1, 0) {
			\small{PBNS \cite{bertiche2021pbns}}};			
		\node[right of=image, xshift=6.9cm,yshift=5.1cm] (textD) at (-1, 0) {
			\small{Simulation \cite{narain2012arcsim}}};			

	\end{tikzpicture}

	\caption{Qualitative comparison with state-of-the-art methods. SNUG generalizes well to unseen body shapes and motions, and produces detailed folds and wrinkles. The results of SNUG are, at least, on par with the realism of \textit{supervised} methods that require large datasets \cite{santesteban2019virtualtryon,patel2020tailor}, and close to state-of-the-art \textit{offline physics-based} simulation \cite{narain2012arcsim}.}
	\label{fig:qualitative_comparison}
\end{figure*}
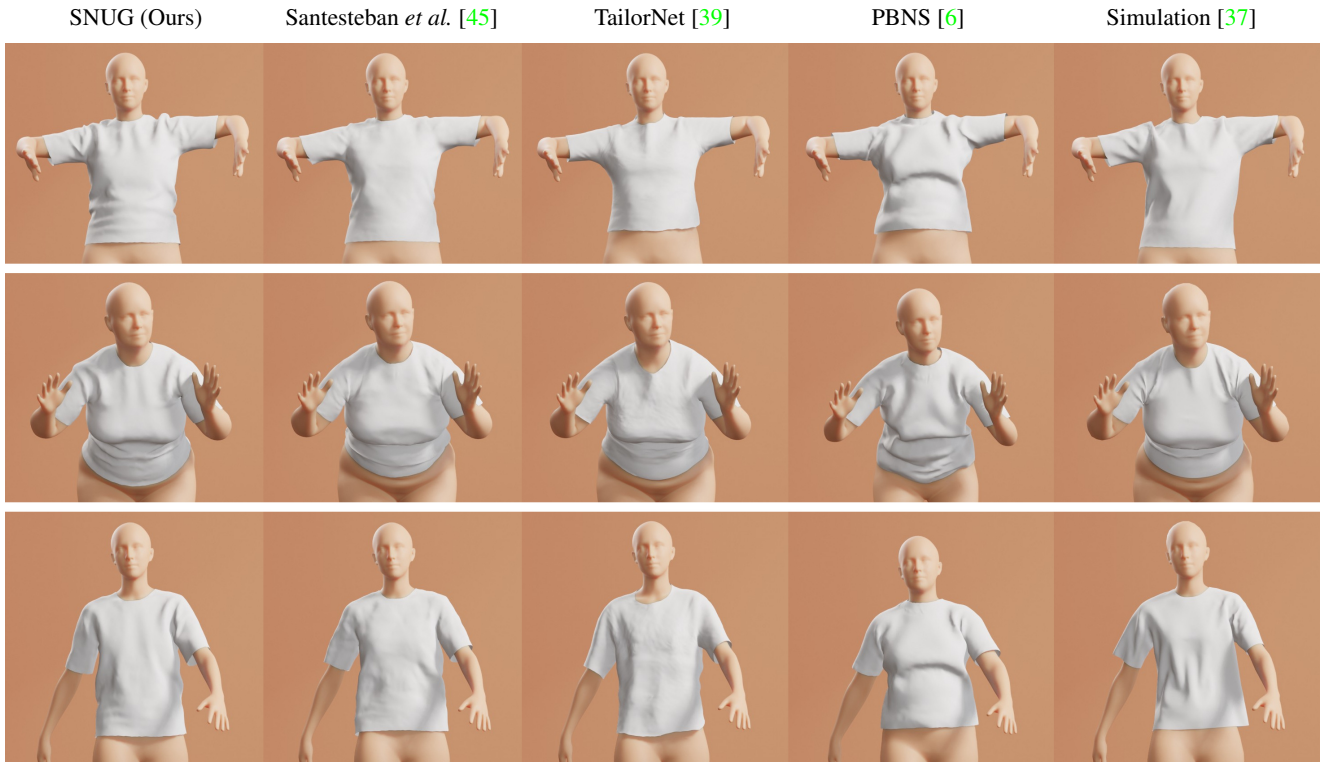

\subsection{Qualitative Evaluation}
We qualitatively evaluate our method in Figure \ref{fig:qualitative_comparison} and, more extensively, in the supplementary video. To this end, notice that we always use body shapes and motions unseen during training.
Additionally, we provide comparisons to the state-of-the-art \textit{supervised} methods of Santesteban \textit{et al.}\cite{santesteban2019virtualtryon} and TailorNet \cite{patel2020tailor}, as well as to the recent work PBNS \cite{bertiche2021pbns} that uses physics-constraints as supervision.
To ease the assessment of the realism of each method, we also show results computed with a physics-based simulator \cite{narain2012arcsim}, but notice that this is a traditional offline method, several orders of magnitude slower.

These results demonstrate that our self-supervised method SNUG produces garment deformations that are, at least, on par with the state-of-the-art \textit{supervised} methods \cite{santesteban2019virtualtryon,patel2020tailor}, while we do not require \textit{any} ground-truth dataset.
For PBNS \cite{bertiche2021pbns}, we use a mean body shape because it does not generalize to different bodies. 
Because PBNS does not model an inertial term and it is limited to a simpler material model, the garment deformations are generally more stiff, less realistic, and do not change naturally as a function of body pose. This is visible in rows 1 and 3 for PBNS in Figure \ref{fig:qualitative_comparison}, where the overall wrinkles are the same despite the significant change in body pose. 

\revised{
To further validate our model, we use the ground-truth simulated dataset (described in Section \ref{sec:training}, used for validation purposes only) to retrain our neural network in a per-vertex supervised manner.
In Figure \ref{fig:supervised} and in the supplementary we qualitatively demonstrate that the self-supervised method learns more detailed wrinkles than the supervised counterpart trained with exactly the same motions.}

Additionally, in Figure  \ref{fig:extra_garments} we show more results for a variety of garments learned with our approach, including t-shirts, tops, sleeveless shirts, pants, and shorts, worn by different body shapes.
Notice how our approach produces different wrinkles for each garment type, pose, and shape combination, demonstrating the generalization capabilities of our self-supervised approach. For this figure, we trained one regressor for each garment type.
In the supplementary video you can see animated results of these garments, showcasing for the first time realistic dynamic deformations of self-supervised learning-based garments.

Finally, in the supplementary video we also show an ablation study of the influence of each term of our loss function. We demonstrate that all terms of Equation \ref{eq:loss} contribute to improve the realism of our predicted garments.

\section{Limitations and Conclusion}
We believe SNUG makes an important step towards efficient learning-based models for 3D garments.
To improve the state-of-the-art, instead of following the standard route of training with more data, adding more explicit supervision, or designing more complex architectures, we show that self-supervision based on physical properties of deformable solids leads to simpler and smaller yet highly-realistic models.

While our physics-based loss terms are the fundamental key to self-supervision, we also want to point out that our strategy of exploiting optimization-based schemes (originally derived for simulation problems) to train a neural network carries a few weaknesses and important considerations to take into account.

Specifically, we notice that the self-supervised network tends to converge to simpler solutions than a traditional simulator. 
For example, although our approach is capable of learning pose- and shape-dependent wrinkles and overall dynamics, we struggle to predict fine-level dynamics.  
We hypothesize that this limitation arises from a fundamental difference in how our method works: while standard simulators solve physics for one frame at a time, our model optimizes thousands of frames simultaneously during training. 
This makes our approach more prone to converge to simpler local minima. Nevertheless, we want to highlight that, despite this limitation, the cloth dynamics learned by our method are on par with other data-driven approaches.

Another aspect open to future research is the collision handling. Although our loss penalizes collisions between the garment and the body in train samples, we found noticeable collisions in test motions. Although these collisions can be efficiently solved with a postprocessing step, we believe it would be valuable to explore ways to enforce this constraint on the network. Addressing self-collisions of the garment is another aspect that would be worth taking into consideration.

\revised{To foster future research on the field, our trained models and the code to run them are available on \url{http://mslab.es/projects/SNUG}.}

\revised{
\vspace{-1cm}
\noindent\paragraph{Acknowledgments.} The work was funded in part by the European Research Council (ERC Consolidator Grant no. 772738 TouchDesign) and Spanish Ministry of Science (RTI2018-098694-B-I00 VizLearning). }

{\small
	\bibliographystyle{ieee_fullname}
	\bibliography{cvpr22}
}

\end{document}